\documentclass[pageno]{jpaper}

\usepackage[normalem]{ulem}
\usepackage{algorithm}
\usepackage{algorithmicx}
\usepackage[noend]{algpseudocode}
\usepackage{mathptmx} 
\usepackage{amsmath}
\usepackage{graphicx}
\usepackage{color}
\usepackage{setspace}
\usepackage[dvipsnames]{xcolor}
\usepackage{mathtools}
\usepackage{soul}
\usepackage{comment}
\usepackage{fancyhdr}
\usepackage{hhline}
\usepackage{array}
\usepackage{xspace}
\usepackage{url}
\usepackage{fancyvrb}
\usepackage{booktabs}
\usepackage{pifont}
\usepackage{flushend}
\usepackage[us,12hr]{datetime}
\usepackage{dblfloatfix}
\usepackage{seqsplit}
\usepackage{multirow}
\usepackage{float}

\usepackage{hyperref}
\usepackage{cite}
\usepackage{subcaption}
\usepackage{caption}

\newcommand{\authspace}[0]{\hspace{60pt}}

\begin{document}
\newcommand\TODO[1]{\noindent{\color{red} {\bf \fbox{TODO}} {\it#1}}}
\newcommand\concision{\noindent{\color{green} {\bf \fbox{Safe!}}}}
\newcommand\methodname[1]{\textsl{{#1}}}
\newcommand\appname[1]{\textsf{{#1}}}
\newcommand\varname[1]{\texttt{{#1}}}
\newcommand\componentname[1]{{{#1}}}
\newcommand{\pluseq}{\mathrel{+}=}
\newcommand{\astar}{A$^\star$ }

\title{Speculative Path Planning}
\author{
    Mohammad Bakhshalipour
    \authspace{}
    Mohamad Qadri
    \authspace{}
    Dominic Guri
}
\date{\centering
    \textit{Carnegie Mellon University}\\%
}

\maketitle


\thispagestyle{empty}

\begin{abstract}
\vspace{1mm}
Parallelization of \astar path planning is mostly limited by the number of possible motions, which is far less than the level of parallelism that modern processors support. In this paper, we go beyond the limitations of traditional parallelism of \astar and propose \textbf{\emph{Speculative Path Planning}} to accelerate the search when there are abundant idle resources. The key idea of our approach is \textbf{\emph{predicting future state expansions relying on patterns among expansions}} and aggressively parallelize the computations of prospective states (i.e. pre-evaluate the expensive collision checking operation of prospective nodes). This method allows us to maintain the same search order as of vanilla \astar and safeguard any optimality guarantees. We evaluate our method on various configurations and show that on a machine with 32 physical cores, our method improves the performance around 11$\times$ and 10$\times$ on average over counterpart single-threaded and multi-threaded implementations respectively. The code to our paper can be found here \href{https://github.com/bakhshalipour/speculative-path-planning}{https://github.com/bakhshalipour/speculative-path-planning}.
\end{abstract}

\section{Introduction}

Graph-based (weighted) \astar search is the cornerstone of many planning algorithms. Given a good heuristic, \astar can significantly reduce the number of state expansions, thereby reducing the search time. In the context of path planning, \astar is widely used in many applications to find a path from a source position to one (or more) destination position(s). \astar evaluates various nodes in the state space and finds an optimal path from the source to the destination. Specifically, whenever a node is expanded, \astar \emph{evaluates its neighbors}, updates its metadata (setting the node to closed), and then expands another node with the highest prospect of being close to the destination. The evaluation of neighbors is essentially done to find out whether the neighbor nodes should be considered for expansion by the algorithm or not. Being \emph{collision-free} is a condition for a node to be added to the open list (a node that is potentially part of a solution), meaning that the position on the map should not cause the robot to collide with an obstacle. The evaluation of this process is called \emph{collision checking}.

Collision checking can be an extremely compute-intensive task, taking up to 99\% of the planning time~\cite{bialkowski2011massively, murray2016microarchitecture}. One solution for improving the execution time is \emph{parallelizing} the collision checking operations. Since the evaluations of neighbors are completely independent, they can be easily parallelized, thereby achieving speedup. However, the number of neighbors is not typically large; in other words, the parallelization degree is severely limited. The number of neighbors is determined by the number of motions the robot can make. For example, on an 8-connected grid, the robot can move in cardinal and intercardinal directions, implying up to eight parallelization operations to pre-compute the status of the immediate neighbours. On the other hand, today's mainstream computing machines support much more parallelism. CPUs with tens or even beyond a hundred threads, and GPUs with thousands of threads are widespread these days. As a result, when running a path planning kernel on a mainstream machine, most of the physical cores remain idle.

\begin{figure}[h]
    \centering
    \includegraphics[width=0.5\textwidth, height=0.1\textwidth]{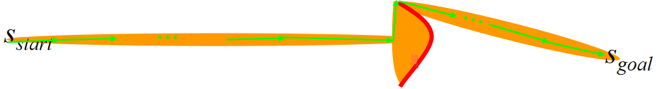}
    \caption{An example of expanded states using weighted \astar.}
    \label{fig:expansion_pattern}
\end{figure}

In this paper, we propose \emph{Speculative Path Planning.} Our objective is to take advantage of idle resources to accelerate the execution time of the $A^*$ and weighted $A^*$ algorithms without breaking their optimality (of $A^*$) and $\epsilon$-suboptimality (of weighted $A^*$) guarantees. We start by making a couple of observations: 1) collision checking is usually an expansive operation that could dominate the search time and 2) given a good heuristic, the search usually follows a pattern of expanding successive nodes along the same directions (for example, the expansions in orange seen in Figure \ref{fig:expansion_pattern}).
 We aim to utilize the available idle threads to perform speculative collision checking of cells likely to be explored in the future to accelerate the graph-search time of the $A^*$ and $weighted-A^*$ algorithms. When a new node $exp\_node$ is expanded using $A^*$ , some (or all) the available threads are first used to find the collision status of its immediate neighbors in parallel. If any available idle threads remain, we use them to compute the collision status of the neighbours of nodes most likely to get expanded in the near future and hence, the search becomes less stalled for performing costly collision checking operations. Importantly, we do not limit our method to compute the collision checking of the current node neighbors; instead, we go further and explore multiple nodes if idle threads are available. These nodes are determined speculatively using a speculation strategy detailed in this report and the collision status of these neighboring nodes is computed in parallel, i.e., we perform speculative parallelization.

We extensively evaluate our approach on a 32-core processor and show that it markedly improves execution time. On four selected 2D maps with 2D planning scenarios, our proposal accelerates execution by 11.1$\times$ over a single-threaded implementation. Compared to a non-speculative parallel implementation that uses an equal number of threads, our approach improves the speedup by 10.4$\times$.

\section{Approach}
\subsection{Speculative path planning}
Algorithm \ref{alg:sp} summarizes the steps of our proposed speculative path planning algorithm. This section will explain each step in detail. 
\begin{algorithm}
  \small
  \begin{algorithmic}[1]
\Procedure{\text{Speculative path planning}}{}
\State $\text{INIT global\_nodes\_state} \gets \emptyset$ 
\State $\text{max\_threads} \gets M $ 
\State $\text{max\_forward\_speculation} \gets S$
\While{!heap.empty()}
    \State $\text{exp\_node} \gets \text{heap.top()}$
    \State $setClosed(\text{exp\_node}) = true$
    \If {\text{exp\_node} == goal}  
    \State $\textit{return path via backtracking}$
    \EndIf
    \State $\text{unkn\_neighs} = \text{getNeighUnknown(\text{exp\_node})}$
    \State $\text{workload\_per\_thread[]} \gets \text{assignWorkload(\text{unkn\_neighs})}$
    \For{$\text{workload in workload\_per\_thread}$} 
        \State $\text{launch\_thread(find\_collision(workload))}$
    \EndFor
    \If {\text{available\_idle\_threads $>$ 0 and $\text{unkn\_neighs} \neq \emptyset$}} 
        \State SPECULATE(\text{exp\_node} )
    \EndIf
    \State join\_threads() 
    \Comment{\textit{Continue with vanilla \astar operations}}
    \For {n in $\text{free\_neighbors}$(exp\_node)}
        \State $n.g \gets cost(\text{exp\_node}) + exp\_node.parent.g$
        \State $n.h \gets \text{get\_heuristic(n)}$
        \If{n.g < previous\_g\_val\_if\_exists(n)}
            \State $\text{add\_to\_open\_list(n)}$
        \EndIf
    \EndFor
\EndWhile
\EndProcedure
  \end{algorithmic}
  \caption{Speculative path planning}
   \label{alg:sp}
\end{algorithm}
\\
We maintain a global map (line 2) indicating the collision status of each node (defined by its $x,y$ position on the map). The collision status can be one of the following:
\begin{itemize}
    \item {\fontfamily{qcr}\selectfont UNKNOWN}: the collision status of the node is unknown and needs to be calculated 
    \item {\fontfamily{qcr}\selectfont FREE}: the node is free.
    \item {\fontfamily{qcr}\selectfont COLLISION}: the node is not free (an obstacle). 

\end{itemize}
We determine the maximum number of threads allocated to the search process (line 4). In line 5, we specify the \textit{max\_forward\_speculation} parameter, which indicates the maximum number of the speculation lookahead searches. Lines 5 to 9 belong to the vanilla $A^*$ algorithm: a new node $exp\_node$ is retrieved from the open list. It's status is then set to closed. If $exp\_node$ is the goal position, we return the final planned path via backtracking.
\subsection{Parallelizing the collision checking operation of immediate neighbors}
Given the newly expanded node $exp\_node$, we have $N$ immediate neighbors (in our experiments N=8 immediate neighbors since we are considering an 8-connected grid). We refer to one of these neighbours as $n_{immediate}$. Either the collision status of $n_{immediate}$ has already been computed (while evaluating previous nodes or using the speculation process) or it is still unknown and needs to be computed. In line 10, we retrieve all of $exp\_node$'s neighbors that have status equal to {\fontfamily{qcr}\selectfont UNKNOWN}. We call this set of nodes $\textit{unkn\_neighs}$. In lines 11-13, we equally split \footnote{Depending on the number of threads available, some threads might be assigned one extra neighbor for computation} the workload of computing collision checking of all nodes in $\textit{unkn\_neighs}$ across the available threads (i.e., each thread could get a list of nodes to compute). In this case, collision checking is performed by what we name "non-speculative threads." If at least a single thread remains, we call the SPECULATE function (lines 14-15). Note that if the set $\textit{unkn\_neighs}$ is empty, we do not perform speculation in order to not stall the main search.

\subsection{Speculation strategy}
\begin{algorithm}
\small
  \begin{algorithmic}[1]
\Procedure{\text{SPECULATE}}{}
        \State \text{dirX = exp\_node.x - exp\_node.parent.x;}
        \State \text{dirY = exp\_node.y - exp\_node.parent.y;}
        \State \text{counter = 1}
        \While{counter $\leq$ \text{max\_forward\_speculation}}
            \State spec\_node.x =  exp\_node.x + counter$\times$dirX
            \State spec\_node.y =  exp\_node.y + counter$\times$dirY
            \For{n in $\text{neigbors(speculated\_node)}$}
                \If {\text{available\_idle\_threads $>$ 0}}
                    \State $\text{launch\_thread(find\_collision(n))}$
                \EndIf
            \EndFor
        \State counter = counter + 1
        \EndWhile
\EndProcedure
  \end{algorithmic}
  \caption{The speculation function}
     \label{alg:speculatefunc}
\end{algorithm}

We converged to a simple speculation strategy that provided a significant increase in search time (Algorithm \ref{alg:speculatefunc}). In lines 2-3, we retrieve the direction $d$ that was taken to reach the currently expanded node $exp\_node$ from its parent node. We then explore nodes in this same direction $d$ up to the maximum lookahead speculation threshold (lines 6-7), which was defined in the parameter \textit{max\_forward\_speculation}. If an explored node $p$  along this path has a status equal to {\fontfamily{qcr}\selectfont UNKNOWN}, we launch a thread to calculate the collision status for \textbf{one of its neighbors only} (line 10). We name these threads \textit{speculative threads}. We compare this to section 2.2 where a thread could potentially calculate the collision status of multiple nodes. By limiting each speculative thread to processing a single neighbor, we are assured that we are never bottlenecked by the speculation process and that we are off the critical path. 
\newline 
Now going back to Algorithm \ref{alg:sp}, in line 16, we join all threads (speculative and non-speculative) and continue with regular $A^*$ operations in lines 17-21 by adding the eligible free neighbors of $exp\_node$  to the open list.
\newline
Note that we have designed and attempted different speculation strategies. For example, by looking further back in the search history (the ancestors  of $exp\_node$). However, we have found that the strategy described above provided the best performing and stable results across different maps.

\subsection{The optimality of the proposed method}
Note that the proposed algorithm maintains the same order of opening and closing of nodes compared to the regular $A^*$ algorithm. As such, our algorithm returns an optimal solution when using $A^*$ search and an $\epsilon-$suboptimal solution when using weighted $A^*$. 
\section{Evaluation}

\subsection{Experimental Setup}
We evaluate our proposal on a WF node of the Narwhal cluster~\cite{narwhal}, with 32 cores. We consider a 2D state space, an 8-connected grid, and a static target. We conduct the experiments on four selected 2D maps with 2D planning scenarios. As a quick proof-of-concept, we use a long latency dummy function named $\textit{find\_collision}$ (line 13 in Algorithm \ref{alg:sp} and line 10 in Algorithm \ref{alg:speculatefunc}) as the collision checking function . This function decrements a large counter to simulate large collision checking time, updates $global\_nodes\_state$ with the correct status of a node (free or an obstacle), and returns. We set the counter such that each collision checking operations takes around 25ms. We use the Euclidean distance as the search heuristic and set the $\epsilon$ value of the search to 1. We use POSIX thread execution model~\cite{butenhof1997programming} for implementing multi-threaded programs.

\subsection{Execution Time}
Figure~\ref{fig:perf} shows the effect of multi-threading and multi-threading plus speculation on execution time. The execution time is normalized relative to the non-speculative single-threaded execution time.

\begin{figure}[h]
  \begin{subfigure}[b]{0.49\columnwidth}
    \includegraphics[width=\linewidth]{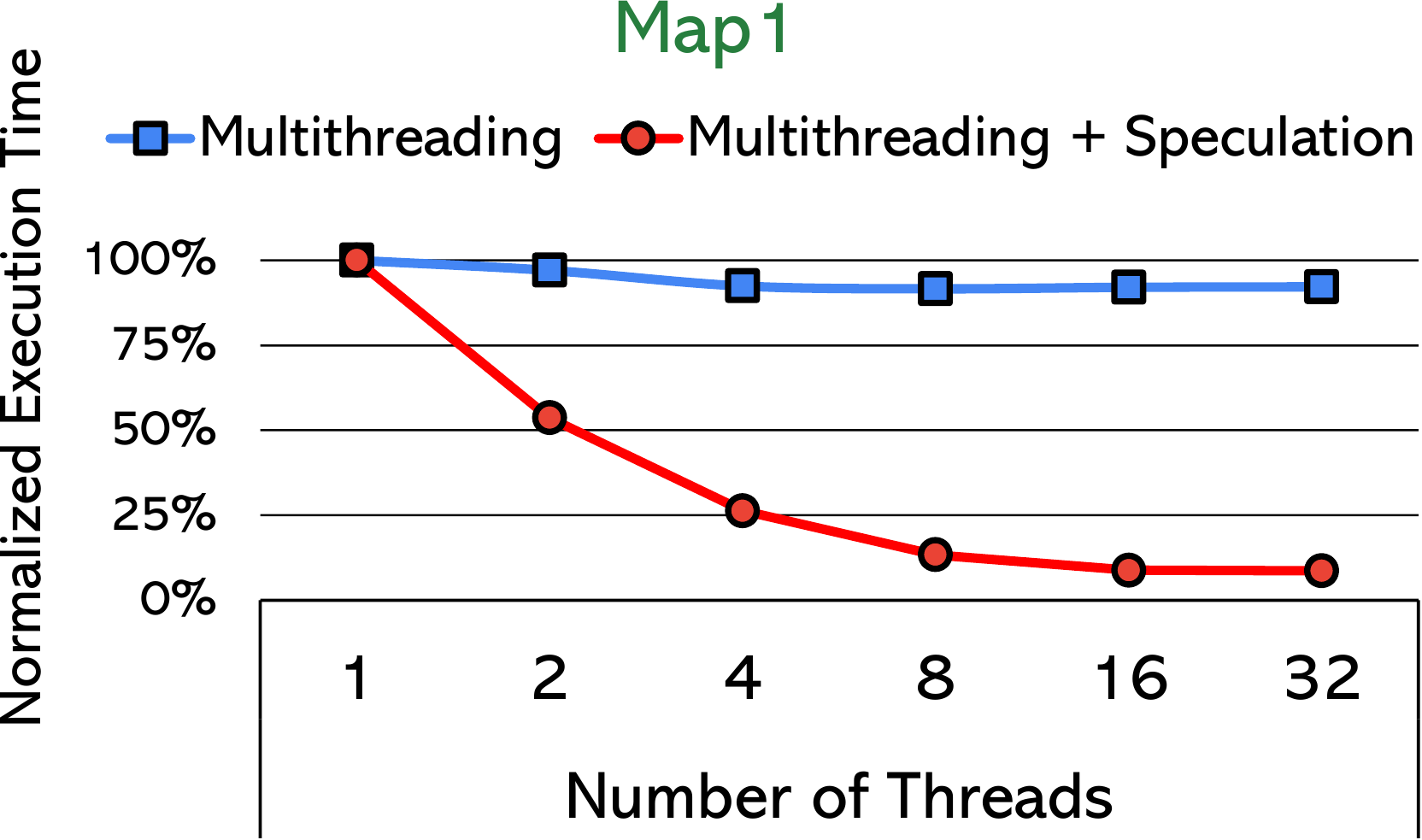}
  \end{subfigure}
  \hfill 
  \begin{subfigure}[b]{0.49\columnwidth}
    \includegraphics[width=\linewidth]{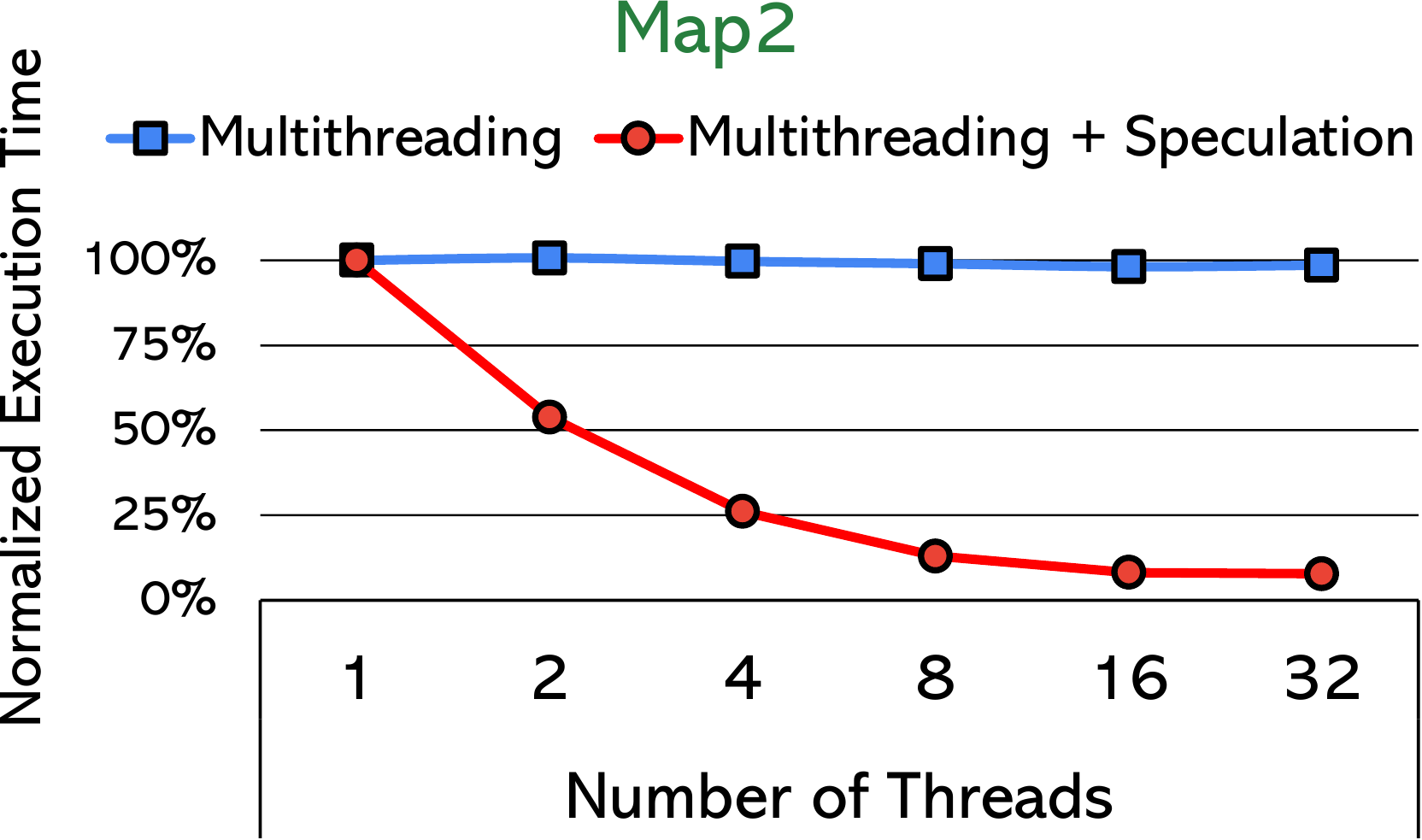}
  \end{subfigure}
  \newline
  \newline
  \begin{subfigure}[b]{0.49\columnwidth}
    \includegraphics[width=\linewidth]{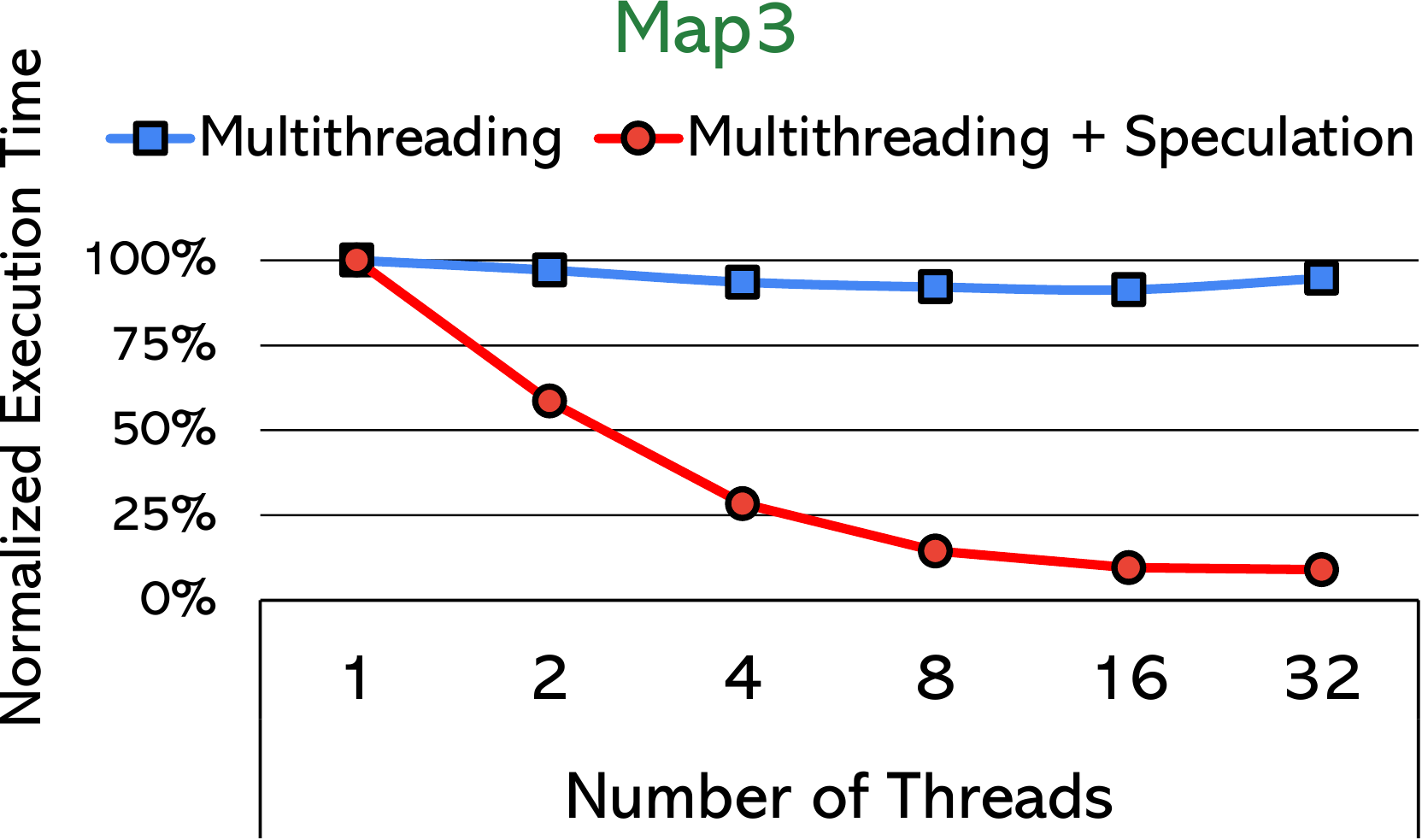}
  \end{subfigure}
  \hfill 
  \begin{subfigure}[b]{0.49\columnwidth}
    \includegraphics[width=\linewidth]{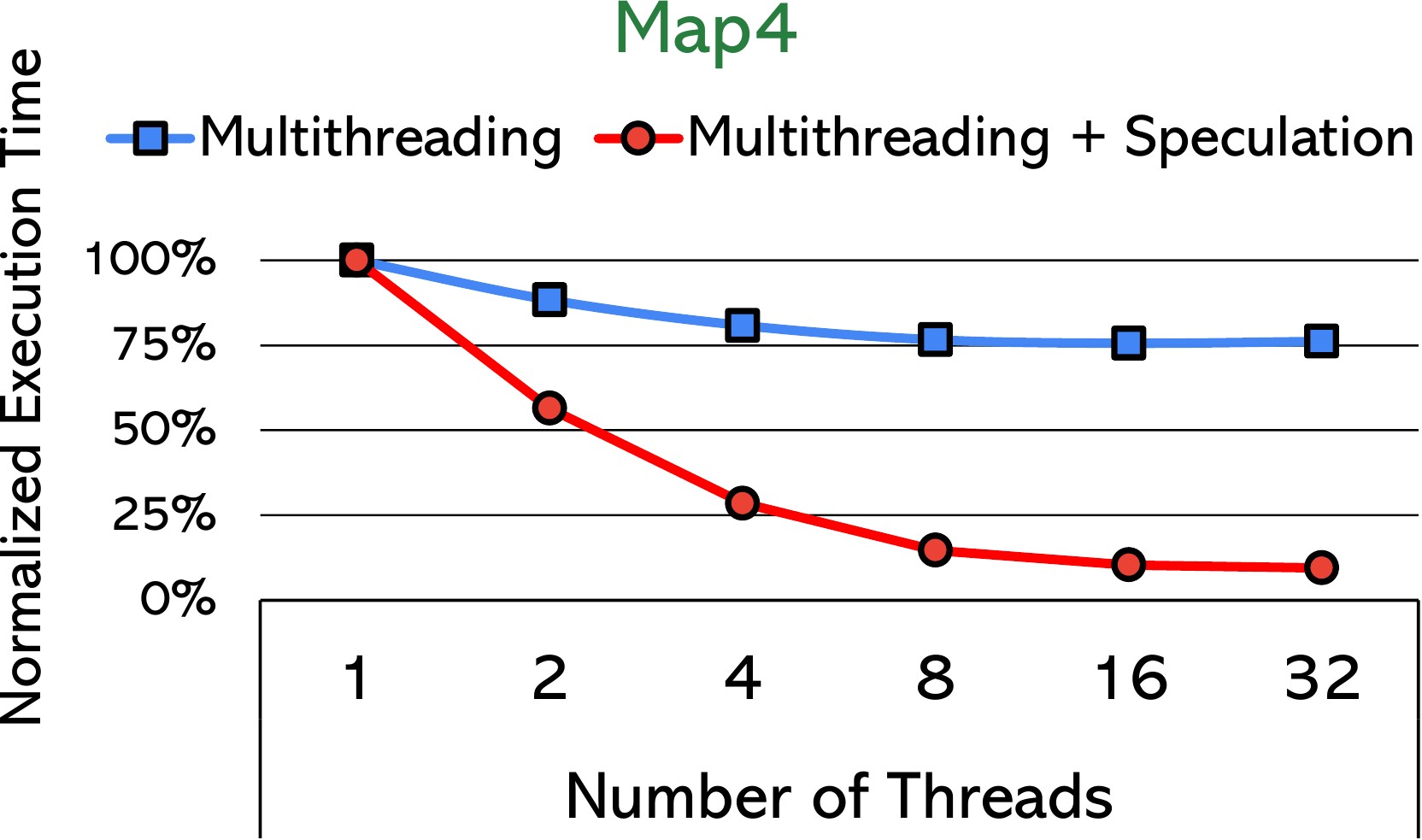}
  \end{subfigure}
    \caption{Execution time with various numbers of threads normalized to a single-threaded implementation.}
    \label{fig:perf}
\end{figure}

Results show that increasing the number of threads with pure multi-threading (i.e., without speculation) gives diminishing returns, and beyond an extent, there is no speedup. Such a limitation is not unexpected since the maximum available parallelism with pure multi-threading equals the number of possible motions (8 in our experiments) and is far less than the maximum available threads (32 in our experiments). However, speculation breaks the parallelization barriers and substantially accelerates the execution time with larger numbers of threads. The average execution time improvement for our approach (with 32 threads) is 11.1$\times$ and 10.4$\times$  relative to non-speculative single-threaded and multi-threaded implementations, respectively.

\subsection{Speculation Accuracy}
\begin{figure}[h]
    \centering
    \includegraphics[width=0.45\textwidth]{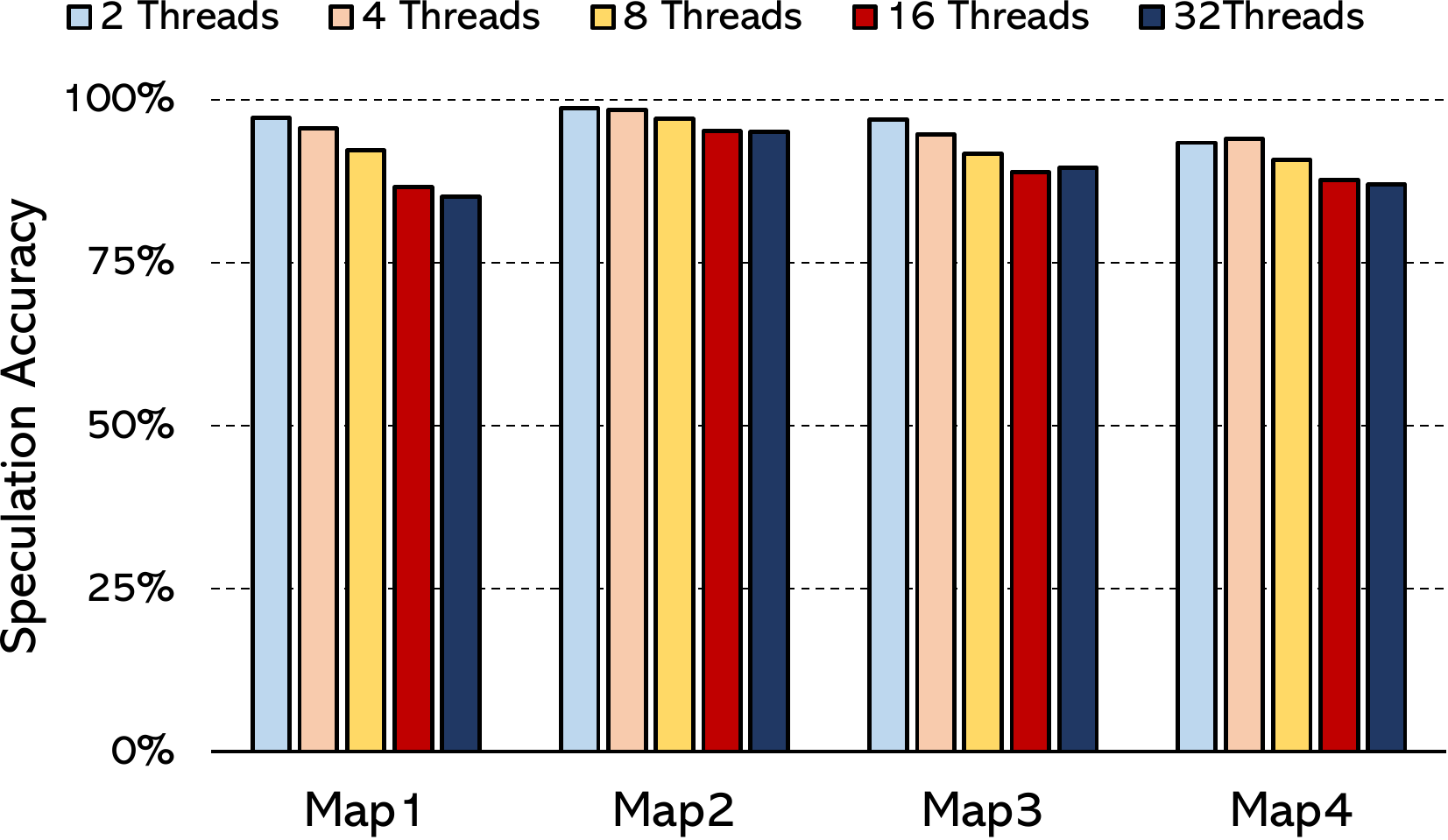}
    \caption{The accuracy of speculations in different configurations.}
    \label{fig:accuracy}
\end{figure}

Figure~\ref{fig:accuracy} shows the \emph{speculation accuracy} of our approach with different numbers of threads on all four maps. We define speculation accuracy as the percentage of speculative computations (i.e., collision checkings) whose result is eventually used by the search. In other words, we consider speculative collision checkings that are performed on nodes that never get used as inaccurate speculations.

As the figure shows, most speculations are accurate, thus substantiating our insight that expansions exhibit specific patterns, leading to their predictability. The average accuracy with 32 threads (worst case) is 89.1\%. For more results on more maps (from \href{https://movingai.com}{https://movingai.com}), see the Appendix. 


\subsection{Division of Labor}

Figure~\ref{fig:dol} shows the average number of collision checking operations per expansion. The figure further shows how many of the operations are performed by non-speculative threads and how many by speculative threads.

\begin{figure}[h]
    \centering
    \includegraphics[width=0.45\textwidth]{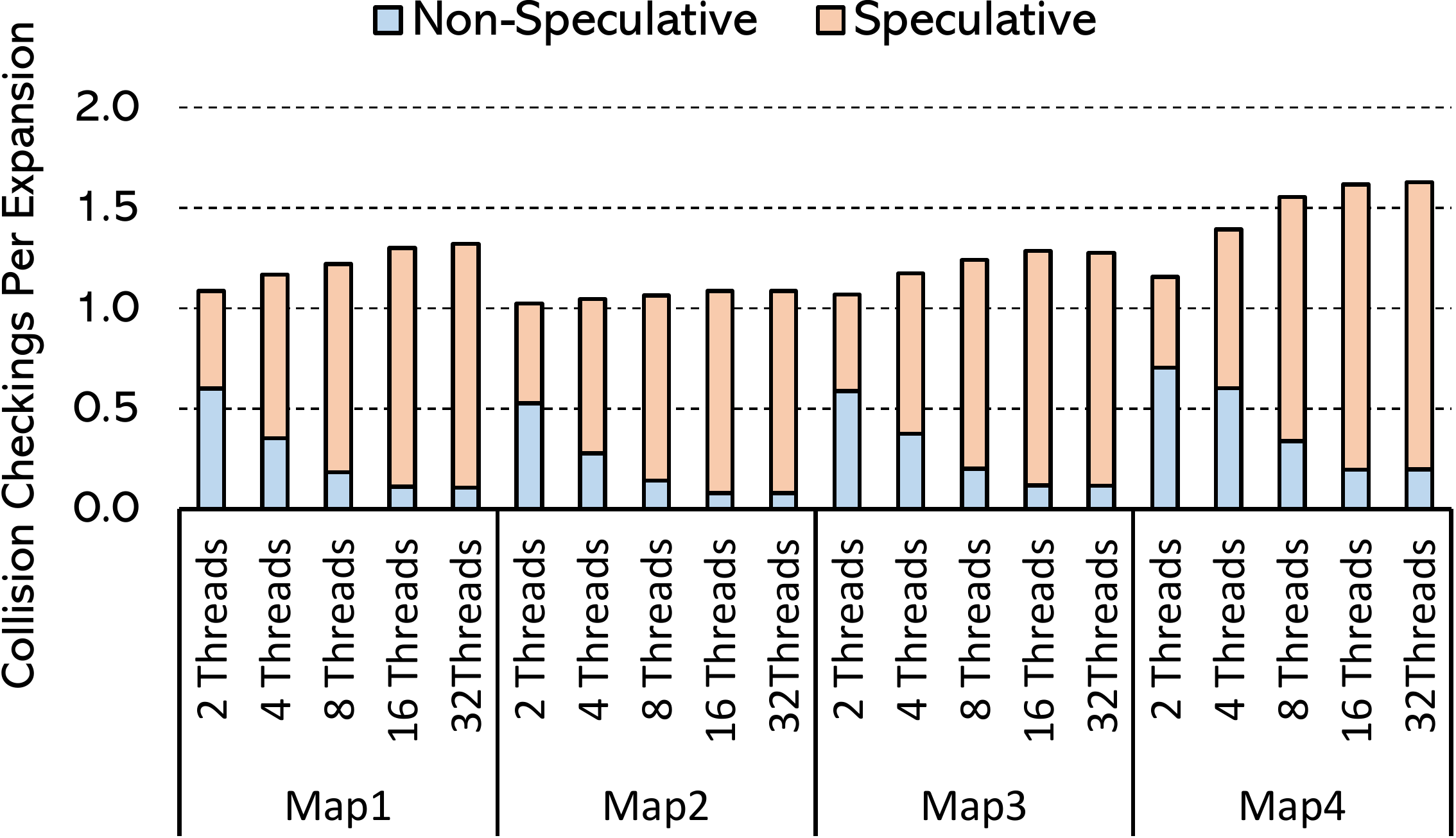}
    \caption{The average number of collision checking operations per node expansion in different setups.}
    \label{fig:dol}
\end{figure}

When increasing the number of threads, the contribution of computations performed by speculative threads goes up. As such, contributions from non-speculative threads decrease; this means that collision checking in the critical path of execution is speculatively evaluated; thus, the program is not stalled by non-speculative collision checking.

\subsection{Demo}

The video\footnote{\href{https://youtu.be/zf6-Sv3IwXg}{https://youtu.be/zf6-Sv3IwXg}} demonstration shows the relative performance of A$^*$ implemented on a single thread versus 16-threads parallel execution with and without speculation. Because the video is generated from log-files, it shows the correct relative progress per frame but not the correct speed. The single-threaded implementation takes 145.9 sec --- comparable to the non-speculative multi-threaded version that took 138.82 sec. Speculation significantly improves the execution time to 18.97 sec.

\section{Conclusion and Future Work}
The \astar algorithm is the backbone of many important planning methods. Accelerating \astar can lead to substantial performance improvements in wide-ranging applications. In this paper, we proposed a method to opportunistically parallelize \astar based on observing that consecutive expansions exhibit recurring patterns. We propose to exploit these patterns by using them to speculatively identify computations that may be required in the future using idle threads; for this paper, this entails running expensive collision avoidance for anticipated future states in a grid-world path planning task. Our approach on a multi-core processor showed that it could accelerate planning by 11.1$\times$ and 10.4$\times$ as compared to vanilla single-threaded and multi-threaded implementations, respectively.

The gist of the approach presented is maximizing computational throughput, which is timely because hardware manufacturers are packing more cores per processor, and newer programming languages like Julia, Swift, and Go are designed to take advantage of such parallelization opportunities. Therefore, future work for this paper involves evaluating our proposal on other substrates like GPUs and FPGAs that offer parallelization opportunities far beyond CPUs. Additionally, we may evaluate the effectiveness of the speculative computation on settings that include realistic collision checking methods, different motion primitives, various heuristic functions, and other graph search scenarios like symbolic planning and task and motion planning (TAMP). 

\bibliographystyle{plain}
\bibliography{pretty_references}

\newpage\phantom{blabla}
\vspace{-10mm}
\section*{Appendix}\label{sec:appendix}

\begin{figure}[!h]
  \begin{subfigure}[b]{\columnwidth}
      \centering
    \includegraphics[width=0.5\textwidth]{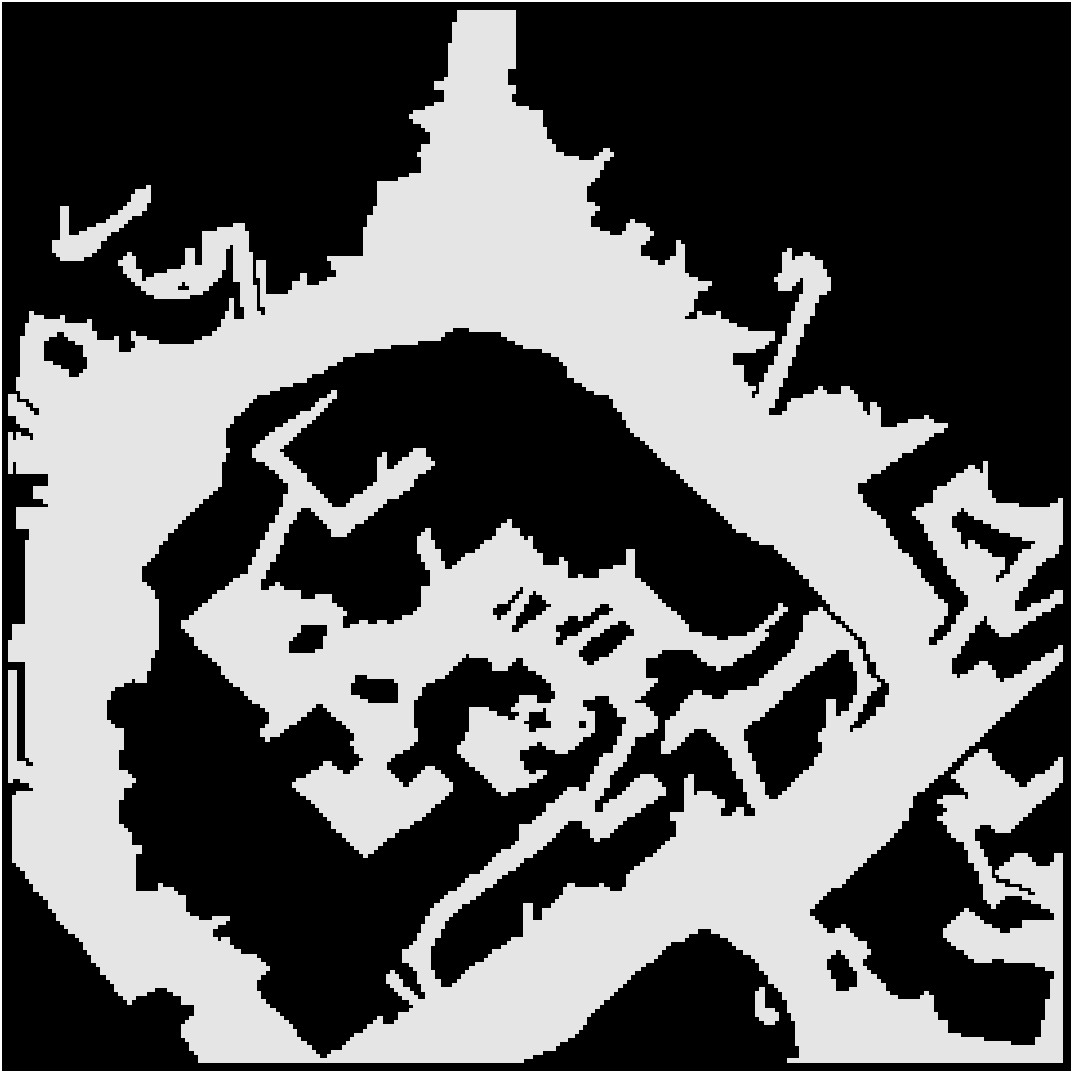}
  \end{subfigure}
  \hfill 
  \begin{subfigure}[b]{\columnwidth}
      \centering
    \includegraphics[width=0.5\textwidth]{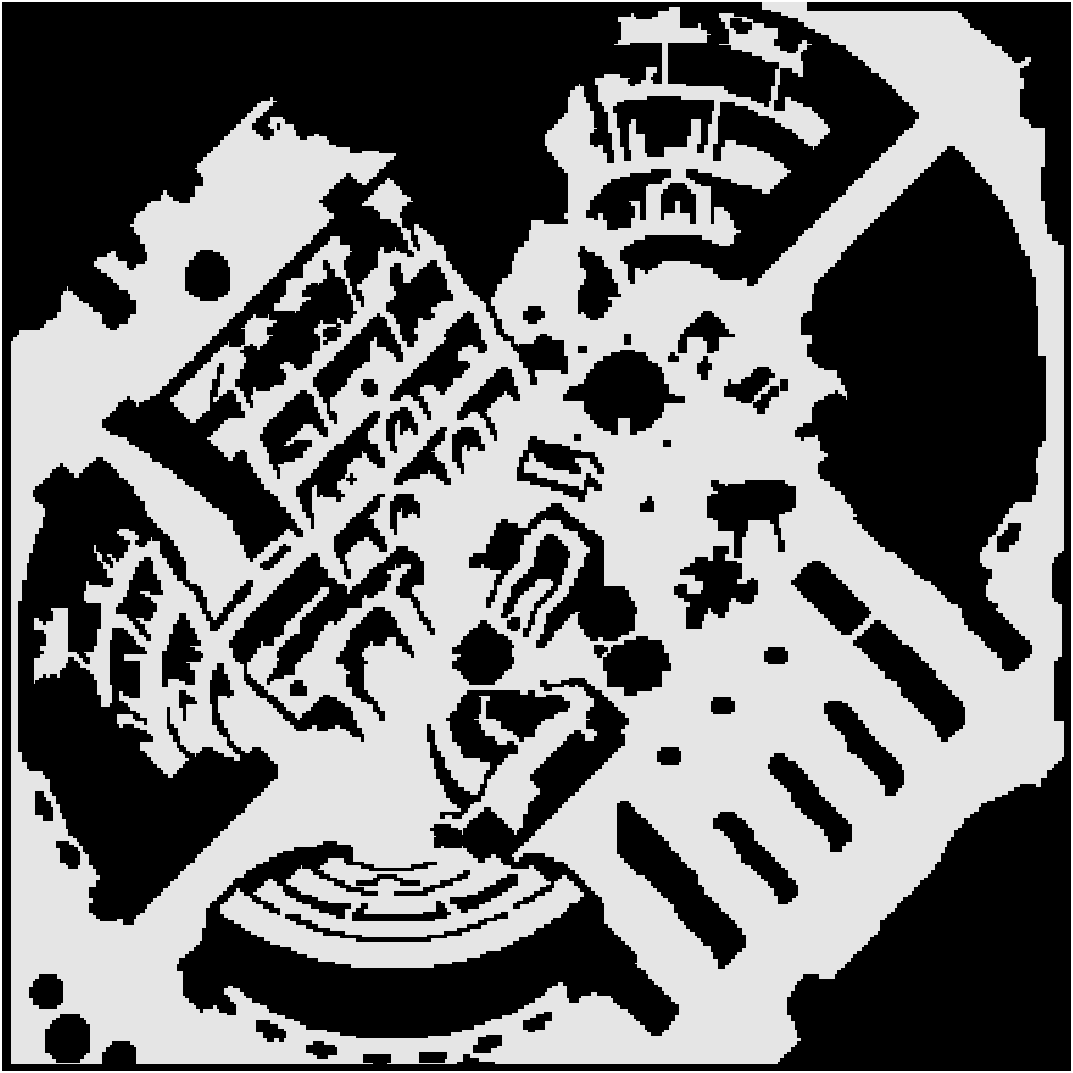}
  \end{subfigure}
  \newline
  \newline
  \begin{subfigure}[b]{\columnwidth}
      \centering
    \includegraphics[width=0.5\textwidth]{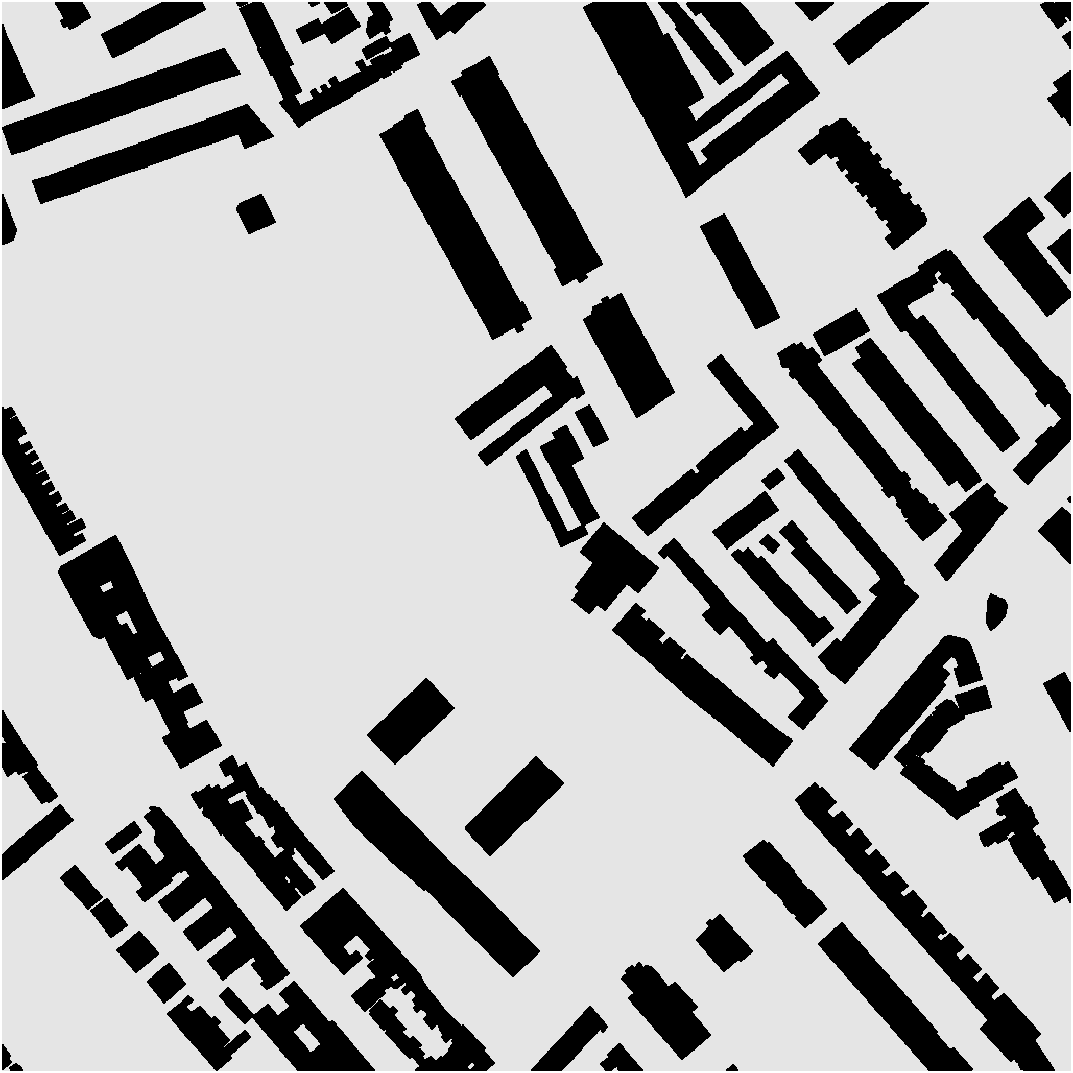}
  \end{subfigure}
  \hfill 
  \begin{subfigure}[b]{\columnwidth}
      \centering
    \includegraphics[width=0.5\textwidth]{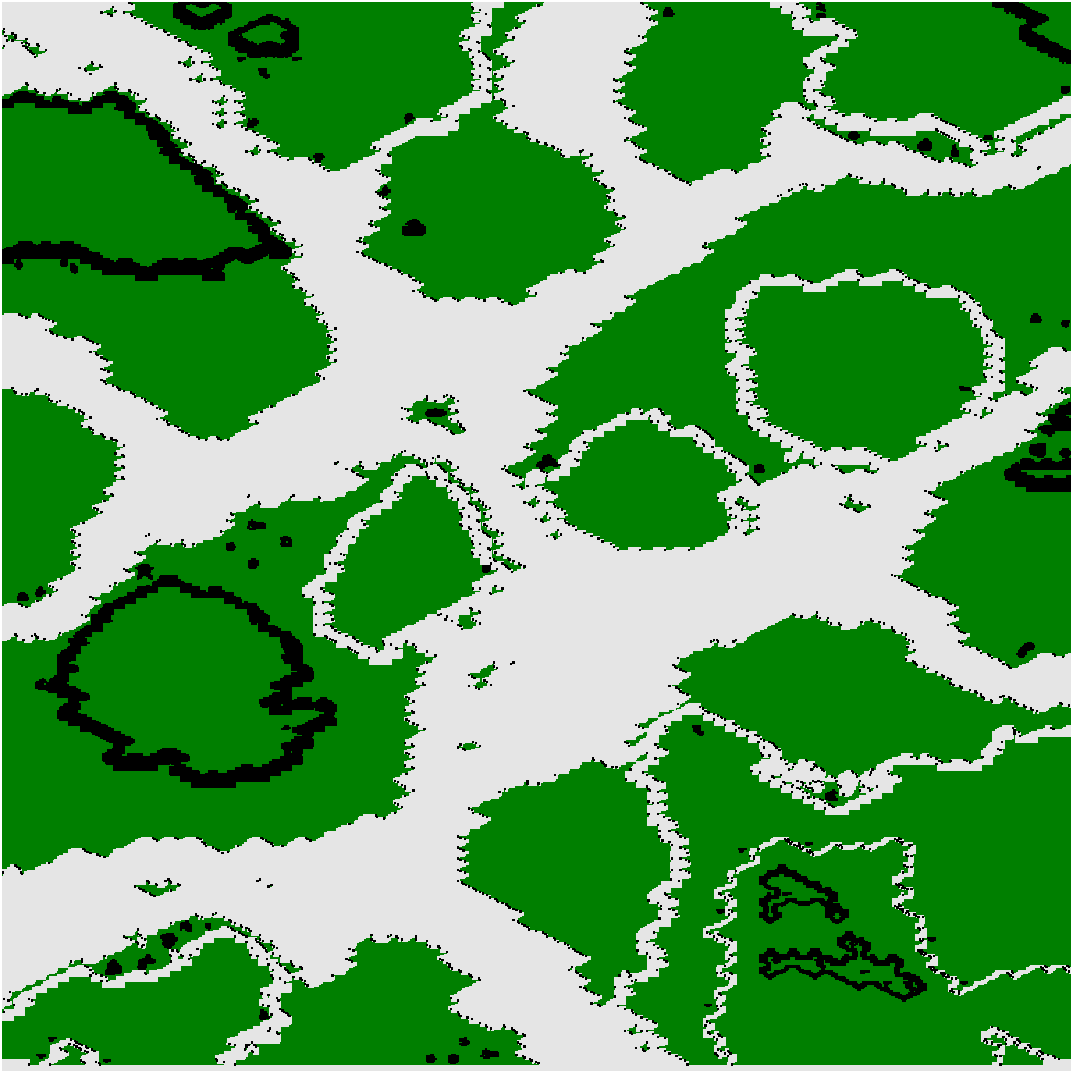}
  \end{subfigure}
    \caption{Experimental maps obtained from \href{https://movingai.com}{https://movingai.com}: AR0400SR, AR0700SR, London, and Sapphire Isles}
    \label{fig:movingai-maps}
\end{figure}

To further test the performance our proposed method we added more maps from \href{https://movingai.com}{https://movingai.com} showing significant execution time reduction with speculation , see Figure~\ref{fig:movingai-maps} .

\begin{figure}[!h]
  \vspace{20mm}
  \begin{subfigure}[b]{\columnwidth}
    \includegraphics[width=0.8\textwidth]{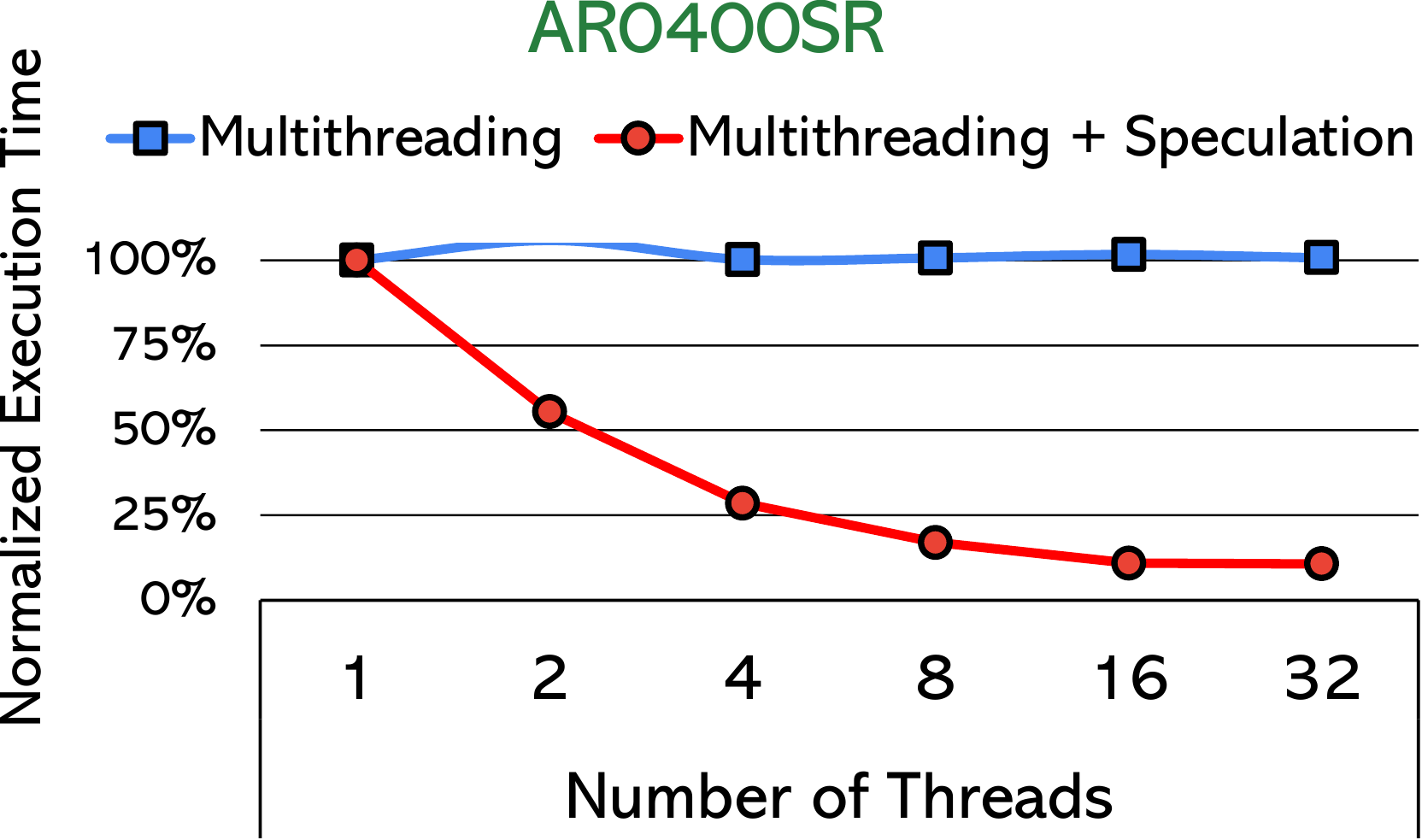}
  \end{subfigure}
  \hfill 
  \begin{subfigure}[b]{\columnwidth}
    \includegraphics[width=0.8\textwidth]{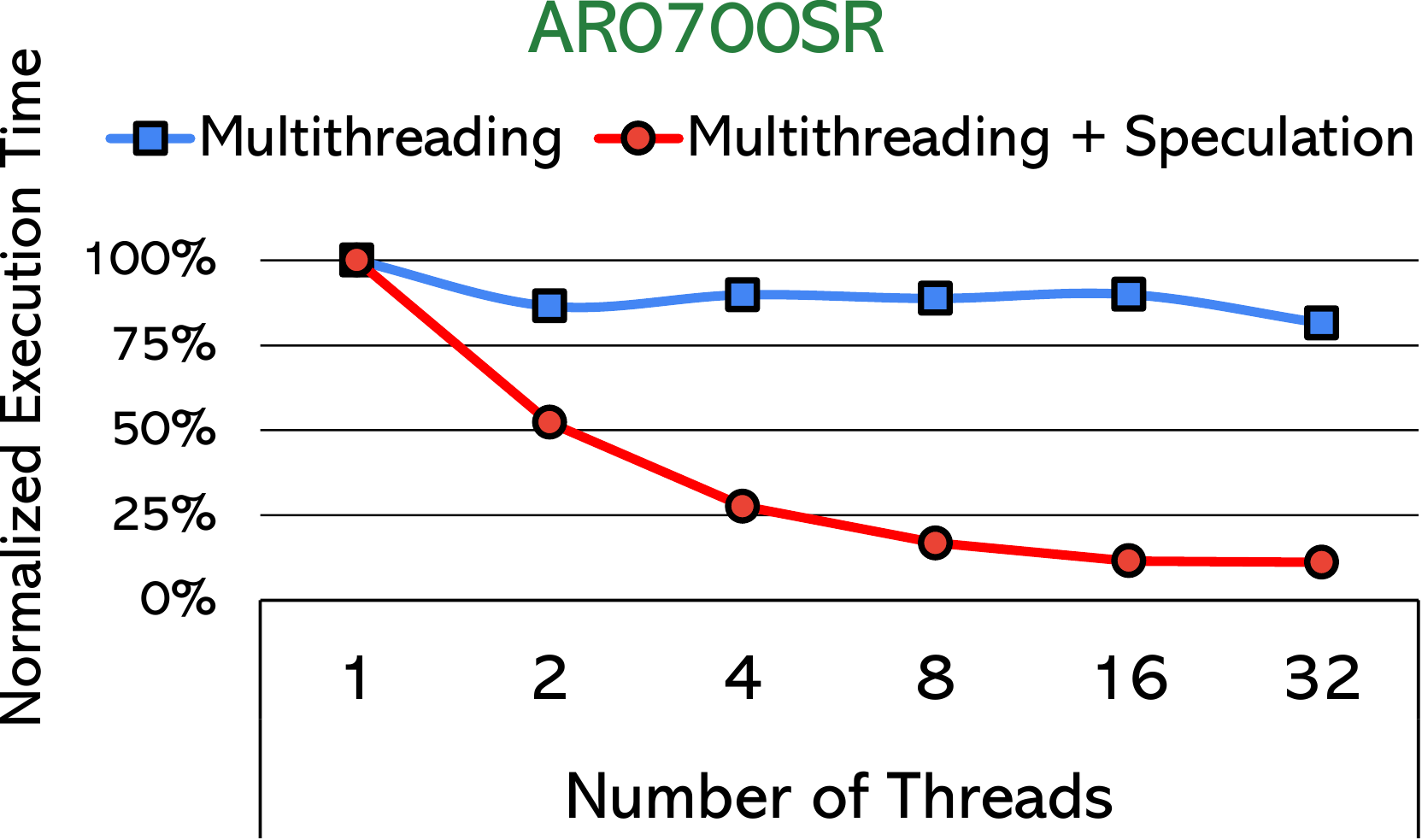}
  \end{subfigure}
  \newline
  \newline
  \begin{subfigure}[b]{\columnwidth}
    \includegraphics[width=0.8\textwidth]{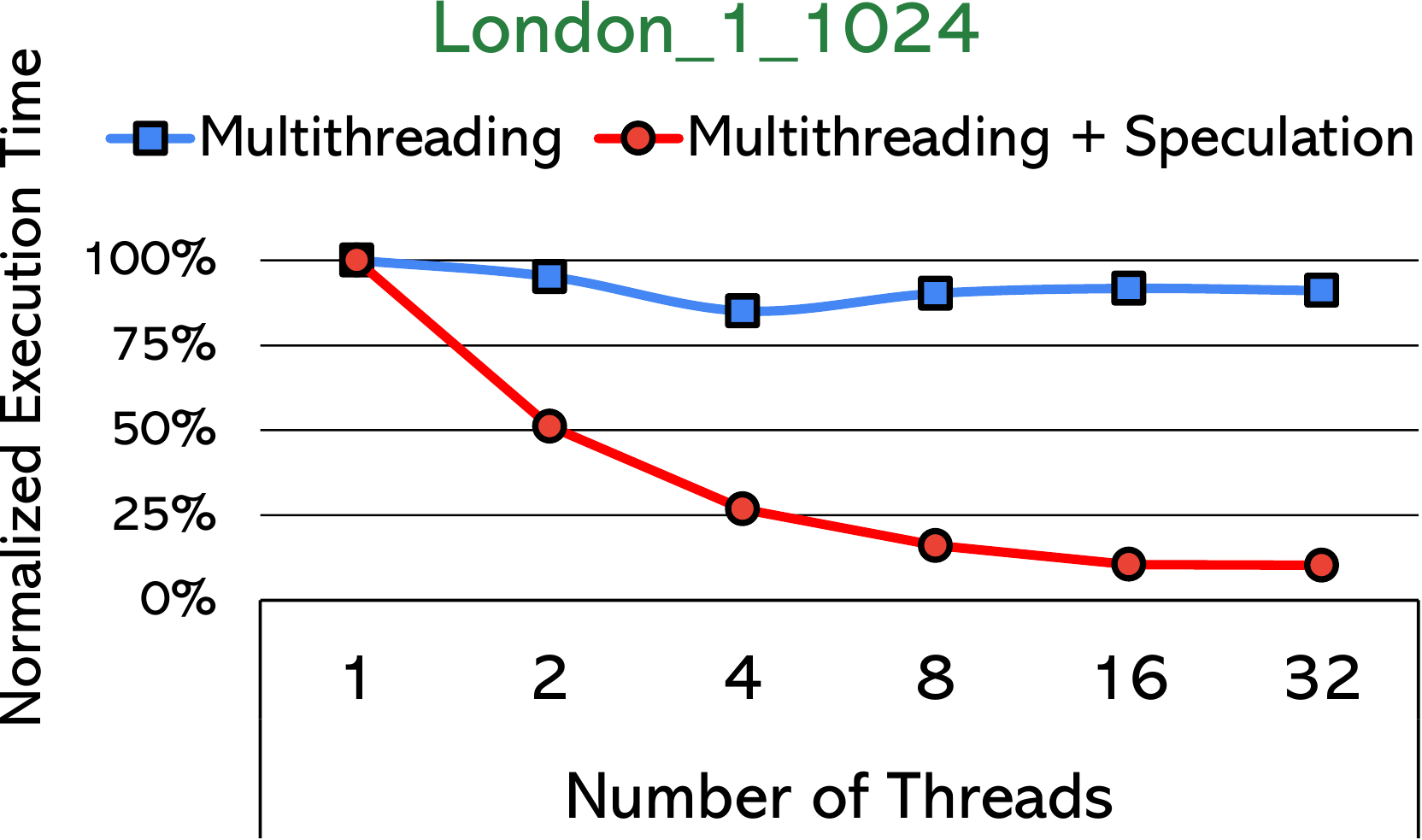}
  \end{subfigure}
  \hfill 
  \begin{subfigure}[b]{\columnwidth}
    \includegraphics[width=0.8\textwidth]{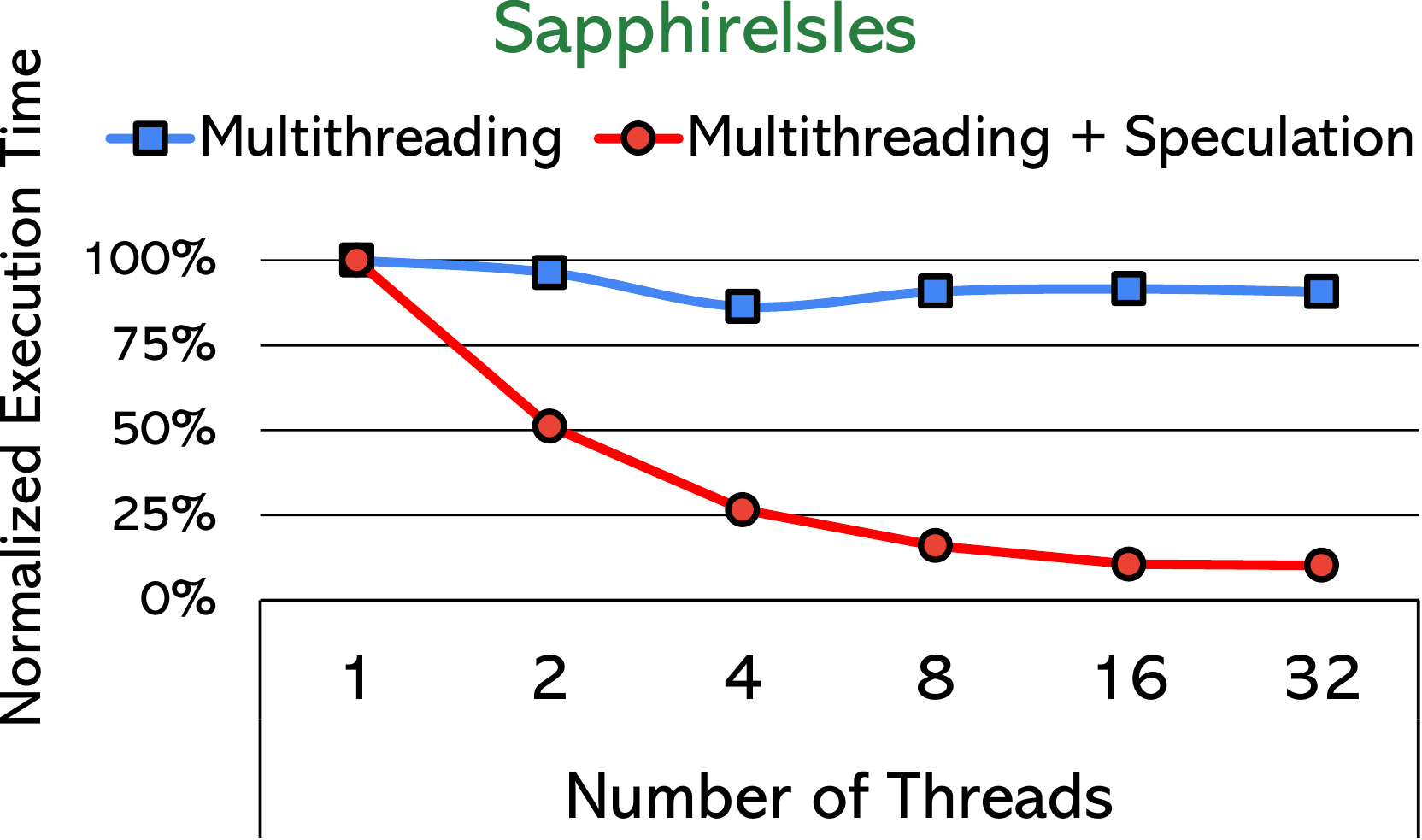}
  \end{subfigure}
    \caption{Execution time with various numbers of threads normalized to a single-threaded implementation for
    \href{https://movingai.com}{https://movingai.com} maps}
    \label{fig:perf}
\end{figure}

\newpage



\newpage

\begin{figure*}[t]
  \vspace{-18cm}
  \begin{subfigure}[t]{\columnwidth}
    \includegraphics[width=0.8\textwidth]{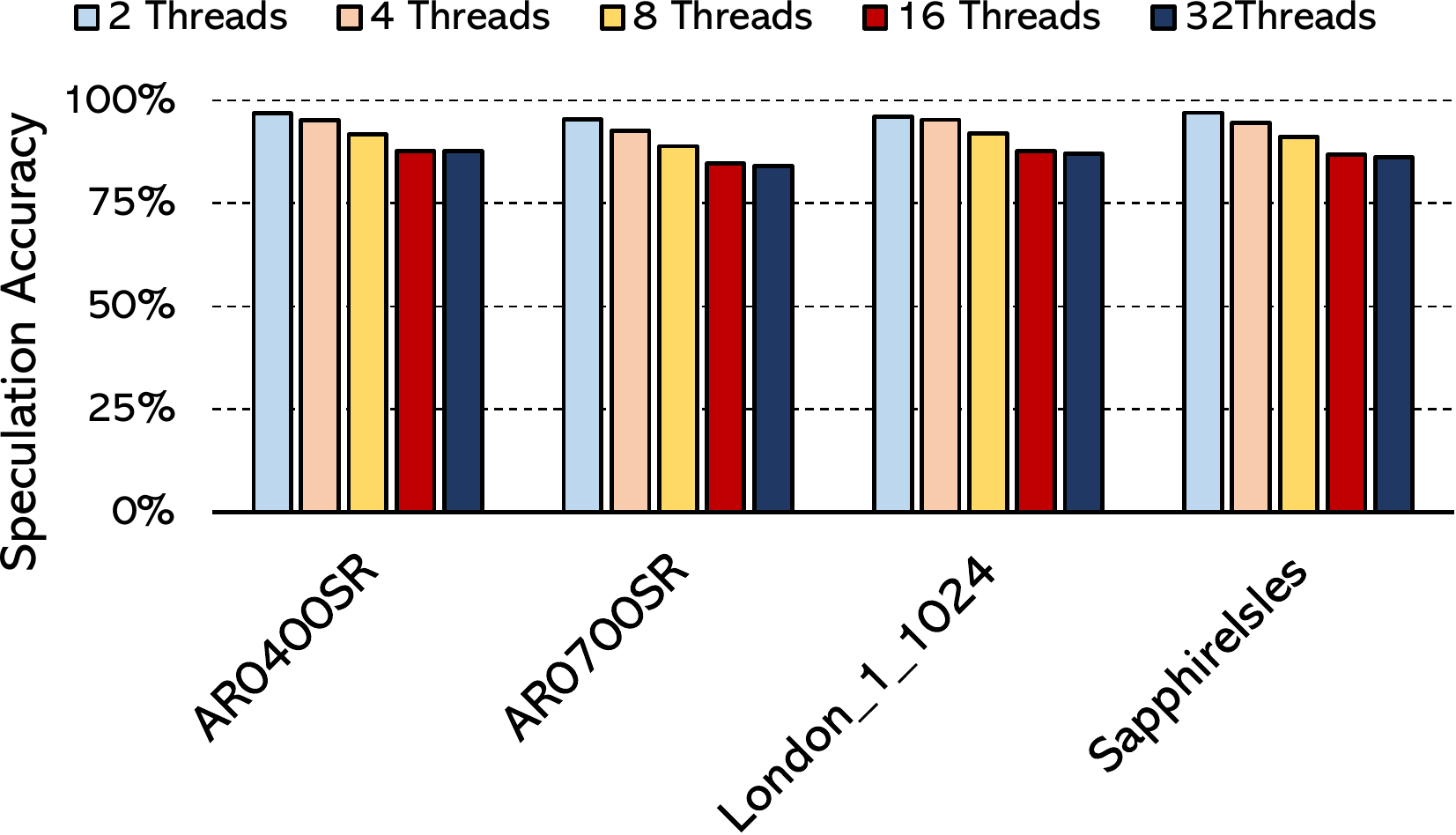}
    \caption{The accuracy of speculations in different configurations.}
  \end{subfigure}
  \hfill 
  \begin{subfigure}[t]{\columnwidth}
    \includegraphics[width=0.8\textwidth]{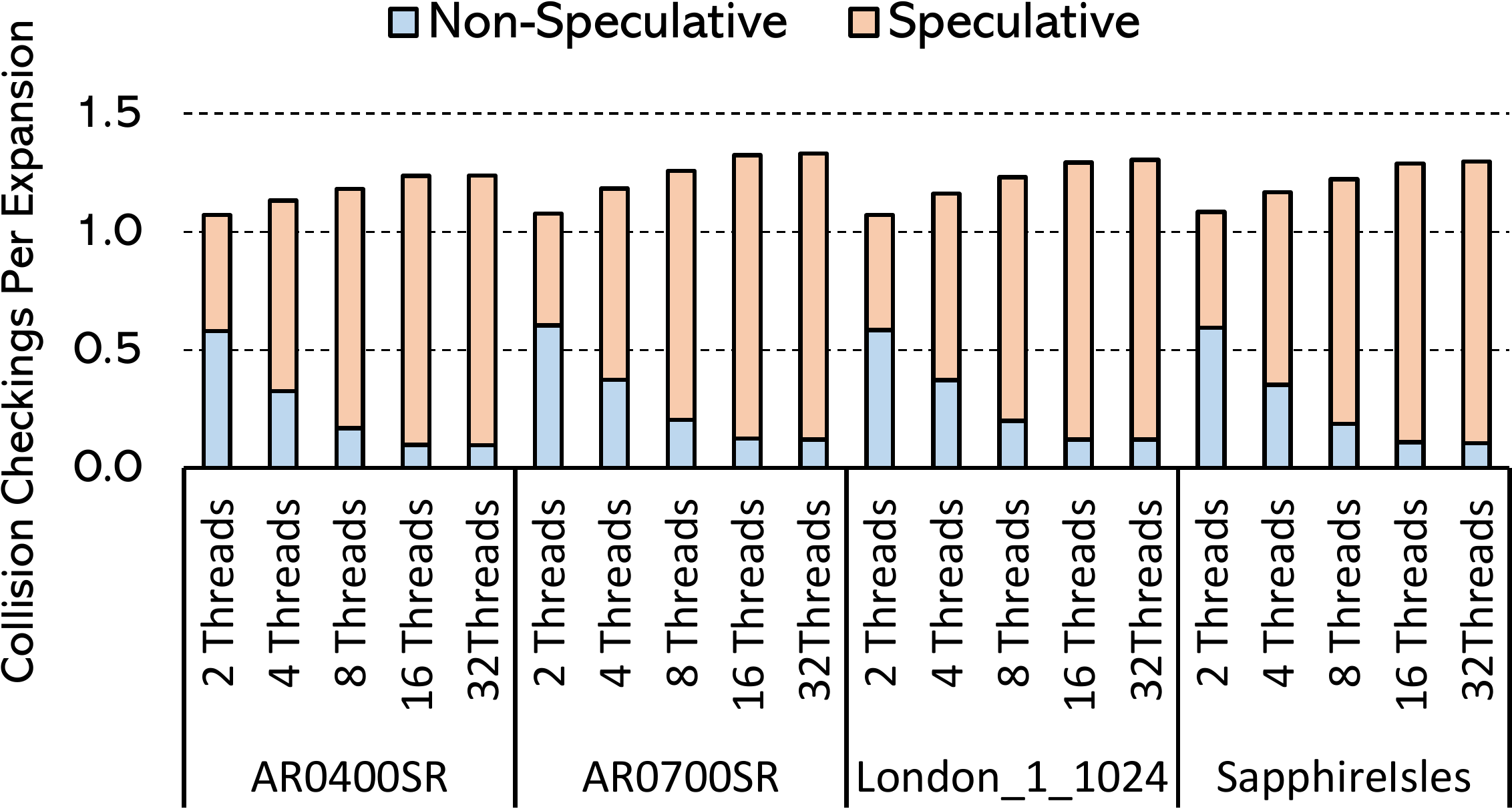}
    \caption{The average number of collision checking operations per node expansion in different setups.}
  \end{subfigure}
  \newline
    \caption{\href{https://movingai.com}{https://movingai.com} maps --- speculation accuracy and collision checkings per expansion.}
    \label{fig:perf}
\end{figure*}

\end{document}